\documentclass[11pt,a4paper,twocolumn]{article}

\usepackage[dvipdfmx]{graphicx, color}
\usepackage{amssymb,amsmath,subfigure}
\usepackage{graphicx,algorithm, algorithmic}

\usepackage{url}
\urldef{\mailsa}\path|{alfred.hofmann, ursula.barth, ingrid.haas, frank.holzwarth,|
\urldef{\mailsb}\path|anna.kramer, leonie.kunz, christine.reiss, nicole.sator,|
\urldef{\mailsc}\path|erika.siebert-cole, peter.strasser, lncs}@springer.com|

\newcommand{\udl}[1]{\underline{#1}}

\newcommand{\changed}[1]{} % これは消してくれる

\begin{document}

\title{Predicting Glaucoma Visual Field Loss \\
by Hierarchically Aggregating Clustering-based Predictors}

\author{
Motohide Higaki$^1$, Kai Morino$^1$, Hiroshi Murata$^2$, 
Ryo Asaoka$^2$, Kenji Yamanishi$^{1,3}$\vspace{0.3cm} \\ 
\small $^1$ Graduate School of Information Science and Technology, The University of Tokyo, Tokyo 113-8656, Japan.\\
\small $^2$ Department of Ophthalmology, The University of Tokyo, Tokyo 113-8655, Japan.\\
\small $^3$ CREST, JST, Honcho, Kawaguchi, Saitama 332-0012, Japan.
}

\maketitle

\begin{abstract}
This study addresses the issue of predicting the glaucomatous visual field loss from patient disease datasets.
Our goal is to accurately predict the progress of the disease in individual patients.
As very few measurements are available for each patient, 
it is difficult to produce good predictors for individuals.
A recently proposed clustering-based method enhances the power of prediction using patient data with similar spatiotemporal patterns.
Each patient is categorized into a cluster of patients, 
and a predictive model is constructed using all of the data in the class.
Predictions are highly dependent on the quality of clustering, but it is difficult to identify the best clustering method.
Thus, we propose a method for aggregating cluster-based predictors to obtain better prediction accuracy than from a single cluster-based prediction.
Further, the method shows very high performances by hierarchically aggregating 
experts generated from several cluster-based methods.
We use real datasets to demonstrate that our method performs significantly better than conventional clustering-based and patient-wise regression methods,
because the hierarchical aggregating strategy has a mechanism whereby good predictors in a small community can thrive.\\
\textbf{keywords: 
glaucoma, hierarchical aggregating strategy}

\end{abstract}

\section{Introduction}

\subsection{Motivation and Purpose of this Study}

In this study, we address the issue of predicting the glaucomatous visual field loss based on patient datasets.
Glaucoma is an eye disease that eventually losses the visual field.
The goal of the present study is 
to accurately predict the glaucoma progression using small visual field datasets.
Because measurements of visual field is expensive, 
early prediction of glaucoma progression based on limited measurements is important for real clinical settings.

A conventional approach for predicting visual field loss is
\textit{patient-wise linear regression} \cite{Holmin:1982} and \textit{clustering-based method} \cite{Liang:2013}.
In patient-wise linear regression, 
we construct a predictor for each individual patient.
However,
as the number of measurements for each patient is very small, 
we cannot produce a good predictor for individuals.
Liang et al. proposed a clustering-based method for glaucoma progression
which utilizes spatiotemporal disease patterns.
Each patient is categorized into a cluster based on each spatiotemporal pattern.
Then, for each cluster, a linear regression predictor is constructed.
A predictor formed from a single cluster is called a \textit{cluster-based predictor};
the clustering-based method selects an appropriate pattern from a pool of cluster-based predictors.
The prediction accuracy of this method is highly dependent on the quality of the clustering method.
Furthermore, a clustering-based method will not work well if the target patient is not a typical member of clusters.

\begin{figure*}[!tb]
 \centering
 \subfigure{\includegraphics[width = 16.0 cm,clip]
{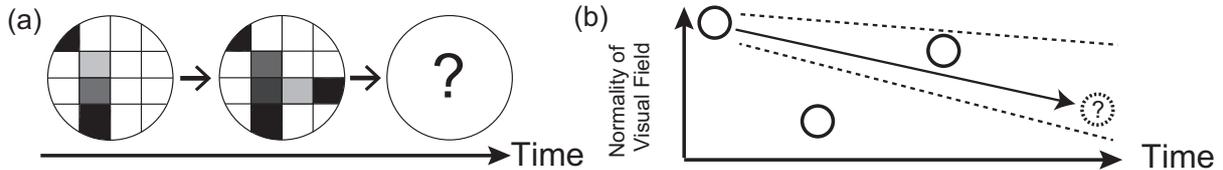}}%
%{./figures/ImageFigure_2014_1118_1116.eps}}%
%{./figures/ImageFigure_2014_0416_0241.eps}}%
 \caption{
Schematic illustration of the prediction of visual fields in glaucoma patients.
(a) Glaucoma progression. 
Regions filled with grayscale colors correspond to local portions of a visual field.
Darkness corresponds to the degree of loss of the visual field.
(b) Schematic of our proposed method. 
Circles show the normality of a visual field at each time point.
The question mark in the dotted circle is the prediction target.
Dotted lines denote cluster-based predictors.
Our proposed method aggregates the cluster-based predictors 
and gives a better prediction.
}
\label{fig:predictingGlaucoma}
\end{figure*}

The aim of the present study is to overcome this weakness in clustering-based methods.
We propose a novel framework for aggregating a number of
cluster-based predictors. 
This significantly improves the prediction accuracy over that of a single cluster-based predictor
in the case of real glaucoma patient datasets.
We present a schematic illustration of our approach in Fig.~\ref{fig:predictingGlaucoma}.

\subsection{Previous Work}

In addition to conventional patient-wise linear regression 
and clustering-based methods \cite{Liang:2013}, 
various techniques have been developed to predict the visual field loss in glaucoma patients.
Bengtsson et al. \cite{Bengtsson:2009} introduced 
the \textit{visual field index} (VFI), 
and showed that only five measurements were needed to produce 
a correlation coefficient of 0.84 with the VFI obtained using all measurements.
Russell et al. \cite{Russell:2012} used Bayesian linear regression to derive 
the \textit{mean sensitivity} index. 
They showed that 
their approach outperformed the ordinary linear regression method 
when there were fewer than nine measurements.
Murata et al. proposed a variational Bayesian learning method \cite{Murata}.
Maya et al. proposed a multi-task learning method 
to predict the visual field loss, 
using matrix decomposition \cite{Maya:2014}.
Maya et al.  also proposed a hierarchical MDL-based clustering method
to detect progressive patterns of glaucoma \cite{MayaKDD}.
Recently, Tomoda et al. proposed a prediction method utilizing information of eye pressure \cite{TomodaHealthinf}.
Other methods were reviewed by Liang et al. \cite{Liang:2013}, 
including those reported by
Noureddin et al. \cite{Noureddin:1991},
Fitzke et al. \cite{Fitzke:1996},
Chan et al. \cite{Chan:2002},  
Chan et al. \cite{Chan:2003},
and Mayama et al. \cite{Mayama:2004}.

However, 
the best predictor  for each patient will differ according to his/her disease features.
In this context, it is crucial to investigate a strategy to aggregate these prediction methods
and automatically to generate the best prediction.
Thus, the present study addresses the issue of combining several prediction methods
to obtain a better prediction than that produced using a single method.
Such aggregating strategies have worked well in other applications, for example,
the issue of predicting prostate specific antigen from short time series \cite{Morino}.

\subsection{Novelty and Significance of this Study}

We propose a new aggregating 
strategy based on
a previously proposed \textit{aggregating algorithm} 
(\cite{Cesa-Bianchi:2006:book}, \cite{vovk:1990}).
In this method, each predictor is treated as an \textit{expert},
and the prediction is obtained 
by taking a weighted average of all experts' predictions.
In theory, 
this method performs almost as well as the best expert in terms of the worst-case regret.
However, in practice,
the aggregating algorithm outperforms all of the experts in terms of the instantaneous prediction loss.

The novelty of the present study is 
a novel aggregating algorithm which is adapted to glaucoma visual loss prediction:
We modify the original aggregating algorithm as follows 
to enable its practical application to 
the specialized setting of glaucoma visual loss prediction:

1) \textit{Learning rate adaptation}: 
The learning rate is estimated using training examples to maximize the improvement rate.
This indicates the improvement in prediction loss
relative to the patient-wise linear prediction.
This learning rate achieves better performance 
than the theoretically designed learning rate.

2) \textit{Batch learning adaptation}: 
The expert weights are determined in a batch process,
rather than the online process
employed in the original aggregating algorithm.
The modified algorithm optimizes the use of all patient data, 
thereby producing better predictions than the original method.

\begin{figure*}[!t]
 \centering
 \subfigure{\includegraphics[width = 16.0cm]
{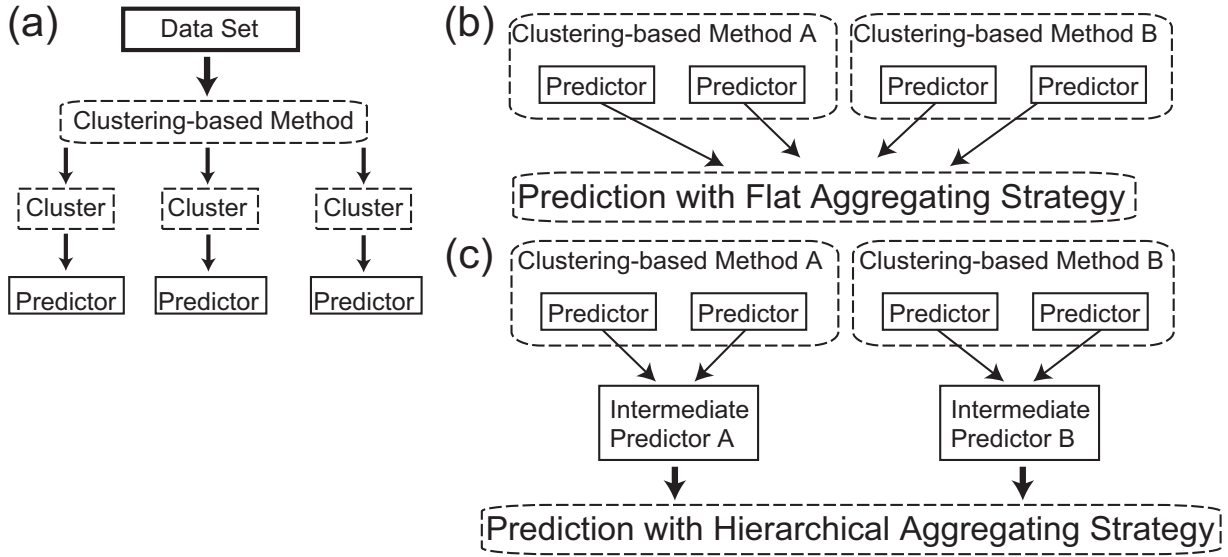}}%
 \caption{
Schematic illustrations of our proposed aggregating algorithm for clustering-based methods.
(a) 
Cluster-based predictors are generated using a clustering-based method.
(b) The flat aggregating strategy.
(c) The hierarchical aggregating strategy.
}
\label{fig:schematic_ill}
\end{figure*}

3) \textit{Hierarchical aggregation}: 
We propose an aggregation process,
the \textit{hierarchical aggregating strategy}.
This method first aggregates all cluster-based predictors 
from a single clustering method 
to obtain intermediate predictors, 
then aggregates these intermediate predictors (see Fig.~\ref{fig:schematic_ill}(c)). 
When we have several clustering-based methods, 
we can aggregate several types of experts.
A facile idea is aggregating all cluster-based predictors 
over all the clustering methods at once (see Fig.~\ref{fig:schematic_ill}(b)).
We call this simple method the \textit{flat aggregating strategy}.
Although the performance of the flat aggregation is better than clustering-based methods,
our analysis indicates that this hierarchical aggregating strategy exhibits 
much better performance than the flat aggregating strategy.

The significance of this study can be summarized as follows:

A) \textit{Better use of predictors in the area of glaucoma visual field loss}: 
We demonstrate that our proposed method delivers better prediction accuracy 
than the flat aggregating strategy, the conventional clustering-based methods, and patient-wise linear regression. 

B) \textit{A new framework for aggregating visual field loss predictors}:
There are a number of promising methods for predicting glaucoma visual field loss, 
and many more are expected to be proposed in the future.
Thus, our method provides a general strategy whereby new methods do not 
have to compete with existing methods. 
Instead, our approach allows numerous methods to collaborate 
in the hierarchical aggregation framework. 
Therefore, by adding new methods to the pool of experts, 
our hierarchical aggregation approach obtains 
better predictions than those given by previous techniques.

The remainder of this paper is organized as follows.
In Section \ref{sec:overview}, we describe the prediction of visual fields.
Section \ref{sec:reviewAA} reviews the aggregating algorithm, 
and Section \ref{sec:generatingexperts} discusses the construction of experts from clustering-based methods.
In Section \ref{sec:algorithm_acm}, 
we present the details of our modification of the existing aggregating algorithm,
and present our experimental results in Section \ref{sec:experiment}.
Section \ref{sec:conclusion} contains our concluding remarks and discussion.

%%%%%%%%%%%%%%%%%%%%%%%%%%%%%%%%%%%%%%%%%%%%%%%%%%%%%%%%%%%%%%%%%%%%%%%%%%%%%%%%%%%%%%%%%%%%%%%%%%%%%%%%%%%%%%%%%%%%%%%%%%%%%
%%%%%%%%%%%%%%%%%%%%%%%%%%%%%%%%%%%%%%%%%%%%%%%%%%%%%%%%%%%%%%%%%%%%%%%%%%%%%%%%%%%%%%%%%%%%%%%%%%%%%%%%%%%%%%%%%%%%%%%%%%%%%
%%%%%%%%%%%%%%%%%%%%%%%%%%%%%%%%%%%%%%%%%%%%%%%%%%%%%%%%%%%%%%%%%%%%%%%%%%%%%%%%%%%%%%%%%%%%%%%%%%%%%%%%%%%%%%%%%%%%%%%%%%%%%
\section{Overview of the Prediction of Visual Fields \label{sec:overview}}

In the present study, the patient visual field comprises 74 scalar real values.
Each value is the \textit{total deviation} (TD), 
which corresponds to the visual loss at each local point of the visual field.
We denote these values in vector form as $\mathbf{y} = (y_1, y_2, \dots, y_D)$,
where $y_i$ ($-30 \le y_i \le 0$) is the TD of the $i$th point  in the visual field,
and $D(=74)$ is the number of points. 
In our algorithm, observations of the target patient's visual field are given 
in this vector form. 
The prediction problem is formalized as follows: 
At time $t$, given past data
$\mathbf{y}_{1},\dots, \mathbf{y}_{t-1}$,
we predict $\mathbf{y}_{t}$ with $\hat{\mathbf{y}}$ using 
 $\mathbf{y}_{1}, \dots, \mathbf{y}_{t-1}$. 
Through a clustering method,
we generate a \textit{cluster-based predictor} from the cluster containing
$\mathbf{y}_{1}, \dots, \mathbf{y}_{t-1}$.
These predictors enables us to produce a prediction using the above vector formalism.

%%%%%%%%%%%%%%%%%%%%%%%%%%%%%%%%%%%%%%%%%%%%%%%%%%%%%%%%%%%%%%%%%%%%%%%%%%%%%%%%%%%%%%%%%%%%%%%%%%%%%%%%%%%%%%%%%%%%%%%%%%%%%
%%%%%%%%%%%%%%%%%%%%%%%%%%%%%%%%%%%%%%%%%%%%%%%%%%%%%%%%%%%%%%%%%%%%%%%%%%%%%%%%%%%%%%%%%%%%%%%%%%%%%%%%%%%%%%%%%%%%%%%%%%%%%
%%%%%%%%%%%%%%%%%%%%%%%%%%%%%%%%%%%%%%%%%%%%%%%%%%%%%%%%%%%%%%%%%%%%%%%%%%%%%%%%%%%%%%%%%%%%%%%%%%%%%%%%%%%%%%%%%%%%%%%%%%%%%
\section{Review of Existing Aggregating Algorithm \label{sec:reviewAA}}

We introduce the basic framework of an existing aggregating algorithm \cite{Cesa-Bianchi:2006:book}.
This method  predicts $\hat{\mathbf{y}}_t$ at time $t$
by combining various \textit{expert} predictions during each time step,  
$\hat{\mathbf{y}}_t = (\sum_{i=1}^N w_{i,t} \mathbf{f}_{i,t})/(\sum_{j=1}^N w_{j,t})$,
where $\mathbf{f}_{i,t}$ is the $i$th expert's prediction at time $t$ and $N$ is the number of experts.
The weights $w_{i,t}$ are updated automatically using the loss function $l(\cdot, \cdot)$ 
of expert prediction $\mathbf{f}_{i,t}$ for outcome $\mathbf{y}_t$ as follows:
\begin{align}
 w_{i,t+1} = w_{i,t} \exp(-\eta l(\mathbf{y}_t, \mathbf{f}_{i,t})), \label{eq:update_online}
\end{align}
where $\eta$ is a constant called the \textit{learning rate}.
The weight assigned to an expert decreases 
when the difference between its prediction and the observation is larger.
This difference is evaluated using loss functions, e.g.,
$l(\mathbf{x}, \mathbf{y}) = (\mathbf{x} - \mathbf{y})^T (\mathbf{x} - \mathbf{y})$.

The performance of this aggregating algorithm is evaluated based on 
the \textit{regret}, 
which is defined as
$\sum_{k=1}^{n} l(\mathbf{y}_k,\hat{\mathbf{y}}_k) - \min_{1\leq i \leq N} \sum_{k=1}^{n} l(\mathbf{y}_k,\mathbf{f}_{i,k})$.
The upper bound of the regret is known to be 
$\ln N/\eta +\eta n/8$ \cite{Cesa-Bianchi:2006:book}.
The optimal learning rate $\eta_* = \sqrt{8\ln N/n}$ is derived by minimizing the regret
to $\sqrt{n \ln N/2}$.
We call this theoretically optimal learning rate $\eta_*$ the \textit{Regret} (RG)-\textit{optimal} $\eta$.
For more details, see \cite{Cesa-Bianchi:2006:book}.

%%%%%%%%%%%%%%%%%%%%%%%%%%%%%%%%%%%%%%%%%%%%%%%%%%%%%%%%%%%%%%%%%%%%%%%%%%%%%%%%%%%%%%%%%%%%%%%%%%%%%%%%%%%%%%%%%%%%%%%%%%%%%%%%%%%%%%%%%%%%%%%%%%%%%%
%%%%%%%%%%%%%%%%%%%%%%%%%%%%%%%%%%%%%%%%%%%%%%%%%%%%%%%%%%%%%%%%%%%%%%%%%%%%%%%%%%%%%%%%%%%%%%%%%%%%%%%%%%%%%%%%%%%%%%%%%%%%%%%%%%%%%%%%%%%%%%%%%%%%%%
%%%%%%%%%%%%%%%%%%%%%%%%%%%%%%%%%%%%%%%%%%%%%%%%%%%%%%%%%%%%%%%%%%%%%%%%%%%%%%%%%%%%%%%%%%%%%%%%%%%%%%%%%%%%%%%%%%%%%%%%%%%%%%%%%%%%%%%%%%%%%%%%%%%%%%
\section{Generating Experts from Clustering Results \label{sec:generatingexperts}}

In this section, we explain how to construct experts from existing clustering methods.
In the process of prediction with a single clustering method,
a cluster-based predictor is generated for each cluster.
The most suitable predictor then gives the prediction.
Such clustering-based methods often produce good predictions,
but it is not clear whether the selected predictor is optimal.
If the target patient is not adequately represented by a single cluster,
none of the cluster-based predictors will produce sufficiently accurate predictions.
In this case, 
combining the contributions from several clusters may improve the prediction accuracy.
Therefore, we can think of cluster-based predictors as experts,
and aggregate them to produce better predictions.
The process of generating experts from clustering-based methods is illustrated in Fig.~\ref{fig:schematic_ill}(a).

\subsection{Spatiotemporal Clustering}

\begin{figure*}[!tb]
 \centering
 \subfigure{\includegraphics[width = 15.0cm]
{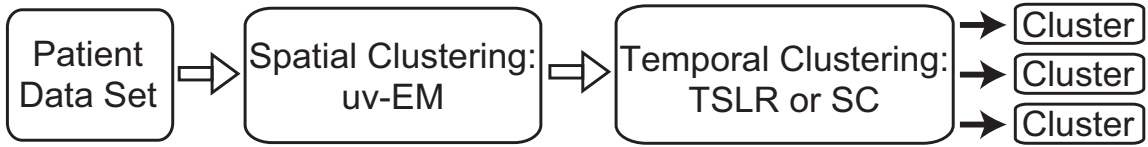}}%
 \caption{
Generating clusters using a spatiotemporal clustering algorithm \cite{Liang:2013}.
}
\label{fig:Spatiotemporal}
\end{figure*}

Liang et al. \cite{Liang:2013} proposed a spatiotemporal clustering algorithm 
that predicts the progress of glaucoma using past patient information.
The algorithm comprises a spatial feature clustering method and
a prediction module that uses the temporal characteristics of each spatial cluster.
This scenario is illustrated in Fig. \ref{fig:Spatiotemporal}.

We employ the spatial feature clustering method (called \textit{uv-EM}) proposed in \cite{Liang:2013}.
This is an \textit{expectation-maximization} (EM) algorithm that learns the spatial cluster centers 
and temporal feature vectors of glaucoma patients' visual fields.

From the clinical knowledge that TD decreases linearly over time,
we assume that the clustering-based predictors can be written as follows \cite{Liang:2013}:
\begin{align}
 \mathbf{y}(t) = \mathbf{w}_1 t + \mathbf{w}_2, \label{eq:linearpredictioncluster}
\end{align}
where $\mathbf{y}(t)$ is the predicted TDs at time $t$, 
and $\mathbf{w}_1$ and $\mathbf{w}_2$ denote the progression rate and 
initial TD, respectively.
These predictors are constructed using the following two methods.
The first method is \textit{temporal-shift linear regression}(\textit{TSLR}), 
where the disease progression rate of all patients belonging to the same spatial cluster 
is assumed to be the same.
However, the initial state of the disease can be different in each patient.
Thus, we obtain a cluster-based predictor through linear regression 
by applying optimal time-shifts to the patients within a cluster.
The second method is \textit{slope clustering}(\textit{SC}), 
where there are at most $C$ progression rates for local visual points in each spatial cluster.

These predictions are more accurate than the traditional patient-wise LR predictors \cite{Liang:2013}, 
which implies that these clustering-based methods represent 
the characteristic of glaucoma patient datasets.
Hereafter, we denote uv-EM + SC as SC and uv-EM + TSLR as TSLR for simplicity
(For more details, see \cite{Liang:2013}).

\subsection{Construction of Experts from Clustering Methods}

Next, we explain how to construct  experts using TSLR, SC, and patient-wise LR.
Each expert is a linear function of time,  as in (\ref{eq:linearpredictioncluster}).
Under TSLR, 
$\mathbf{w}_1$ is as the cluster-based predictor's gradients.
For SC, 
$\mathbf{w}_1$ is the same as that of patients within the training datasets, 
and are selected from $W := \{w^{(1)}, w^{(2)}, \dots, w^{(C)}\}$. 
Patient-wise LR gives $\mathbf{w}_1$ by linearly regressing each patient.

The initial disease state $\mathbf{w}_2$ is determined as
 $\mathbf{w}_2 = \sum_{t=1}^T( \mathbf{y}_t - \mathbf{w}_1 d_t)/T$, 
where $d_t$ is the $t$th date, 
$\mathbf{y}_t$ is the $t$th observation, and $T$ is the number of observations.
This is obtained by minimizing the errors
$E(T) = \sum_{t=1}^T || \mathbf{y}_t - \mathbf{w}_1 d_t - \mathbf{w}_2||_2^2$.
The pair $\mathbf{w}_1$, $\mathbf{w}_2$ defines an expert in the proposed framework.

%%%%%%%%%%%%%%%%%%%%%%%%%%%%%%%%%%%%%%%%%%%%%%%%%%%%%%%%%%%%%%%%%%%%%%%%%%%%%%%%%%%%%%%%%%%%%%%%%%%%%%%%%%%%%%%%%%%%%%%%%%%%%%%%%%%%%%%%%%%%%%%%%%%%%%%%%%%%%%
%%%%%%%%%%%%%%%%%%%%%%%%%%%%%%%%%%%%%%%%%%%%%%%%%%%%%%%%%%%%%%%%%%%%%%%%%%%%%%%%%%%%%%%%%%%%%%%%%%%%%%%%%%%%%%%%%%%%%%%%%%%%%%%%%%%%%%%%%%%%%%%%%%%%%%%%%%%%%%
%%%%%%%%%%%%%%%%%%%%%%%%%%%%%%%%%%%%%%%%%%%%%%%%%%%%%%%%%%%%%%%%%%%%%%%%%%%%%%%%%%%%%%%%%%%%%%%%%%%%%%%%%%%%%%%%%%%%%%%%%%%%%%%%%%%%%%%%%%%%%%%%%%%%%%%%%%%%%%
\section{Algorithm for Aggregating Clustering Methods \label{sec:algorithm_acm}}

To utilize the experts introduced above, 
we modify the original aggregating algorithm as follows:
  1) The learning rate $\eta$ is selected to ensure that it is practical;
  2) the weight is determined in a batch process, rather than an online process;
  3) a hierarchical aggregation 
 algorithm is applied to the clustering-based methods.
The pseudo-code is shown in Alg.~\ref{alg:clusters}.

\begin{algorithm} [!t]
 \caption{Aggregation of clustering methods}
 \label{alg:clusters}
 \begin{algorithmic}
 \STATE \textbf{Input}: Observables $\mathbf{y}_1, \dots, \mathbf{y}_n$ for the target patient.
 \STATE Determine $\mathbf{w}_1$ based on past patient datasets using TSLR, SC, and LR.
 \STATE Determine $\mathbf{w}_2$ from the observations.
 \STATE Determine RG-optimal and IR-optimal values of $\eta$ using the experts.
 \STATE Determine the weights of the experts based on these values of $\eta$.
 \STATE \textbf{Output}: Make prediction using the aggregating algorithm.
 \end{algorithmic}
\end{algorithm}

\subsection{Empirical Determination of the Optimal Learning Rate $\eta$}

As described in Section~\ref{sec:reviewAA}, 
the existing algorithm derives a theoretical RG-optimal $\eta$.
Our goal is to minimize the instantaneous prediction loss at a desired time point;
thus, the regret is not appropriate.
 The \textit{root mean squared error} (\textit{RMSE}) was used to measure the prediction accuracy as
$\sqrt{\sum_{j=1}^{D}(\hat{y}_j-y_j)^2/D}$
where $\hat{\mathbf{y}}=(\hat{y}_1, \dots, \hat{y}_{D})$ are the predictions 
and $\mathbf{y} = (y_1, \dots, y_{D})$ are the observations.
The performance of the prediction method is measured in terms of the 
\textit{Improvement Rate}(\textit{IR}), which is defined as
\begin{equation}
 \mathrm{IR}(n)=\frac{1}{N(n)}\sum_{i=1}^{N(n)}a_i(n), \label{eq:IR_define}
\end{equation}
where $n$ is the number of observation points ($n \in [2, 10]$)
and $N(n)$ is the number of patients in the test dataset.
The value of $a_i(n)$ is 
$1-\dfrac{\mathrm{RMSE_{\mathbf{f}}}(i)}{\mathrm{RMSE_{LR}}(i)}$ if $n < L_i$, and 
$0$ otherwise.
Here, 
$\mathrm{RMSE_{LR}}(i)$ is the RMSE of the $i$th patient using patient-wise LR,
and
$\mathrm{RMSE_{\mathbf{f}}}(i)$ is the RMSE using prediction method ${\mathbf{f}}$.
The larger the value of IR, 
the greater the improvement of ${\mathbf{f}}$ relative to the patient-wise LR.
When ${\mathbf{f}}$ has the same accuracy as the patient-wise LR, IR $=0$.
We empirically determine the 
\textit{improvement-rate} (IR)\textit{-optimal} $\eta$
by maximizing the IR.

\subsection{Determining Expert Weights Using a Batch Process}

In the usual online algorithm, 
the expert weights are updated after each new observation.
For the glaucoma datasets,
samples are usually obtained at intervals of a few months.
Thus, it is natural to use all available data points 
to determine the weights of the experts.
Our method employs the following batch update rule:
$w_{i,t+1} = \exp\Bigl(-\eta \sum_{\tau=1}^t l(\mathbf{y}_\tau, \mathbf{f}_{i,\tau}) \Bigr)$,
which replaces the traditional rule (\ref{eq:update_online}).

\subsection{Hierarchical Aggregation}

We propose a hierarchical aggregating strategy, where
all the predictors given by each single clustering-based method are first aggregated to construct intermediate predictors.
These intermediate predictors are then aggregated to construct the final predictor (see Fig.~\ref{fig:schematic_ill}(c)). 

The purpose of the proposed method is to adequately deal with experts generated from a small expert source.
The number of experts generated from each clustering-based method is naturally different.
In the flat aggregating strategy, 
a good expert belonging to a small community may not greatly affect the prediction 
because a large community gathers considerable expert weights.
On the contrary, 
in the hierarchical aggregating strategy, experts within each community are aggregated to generate intermediate predictors.
Then, the final prediction is given by aggregating intermediate predictors that only one intermediate predictor belongs to each community.
Therefore, such effects on the biased community size may be canceled at the second aggregation.
In the real glaucoma datasets,
the number of experts generated from clustering-based methods is widely distributed,
therefore, our proposed method showed the best performance among all the methods as shown in the next Section~\ref{sec:experiment}.

%%%%%%%%%%%%%%%%%%%%%%%%%%%%%%%%%%%%%%%%%%%%%%%%%%%%%%%%%%%%%%%%%%%%%%%%%%%%%%%%%%%%%%%%%%%%%%%%%%%%%%%%%%%%%%%%%%%%%%%%%%%%%%%%%%%%%%%%%%%%%%%%%%%%%%%%%%%%%%
%%%%%%%%%%%%%%%%%%%%%%%%%%%%%%%%%%%%%%%%%%%%%%%%%%%%%%%%%%%%%%%%%%%%%%%%%%%%%%%%%%%%%%%%%%%%%%%%%%%%%%%%%%%%%%%%%%%%%%%%%%%%%%%%%%%%%%%%%%%%%%%%%%%%%%%%%%%%%%
%%%%%%%%%%%%%%%%%%%%%%%%%%%%%%%%%%%%%%%%%%%%%%%%%%%%%%%%%%%%%%%%%%%%%%%%%%%%%%%%%%%%%%%%%%%%%%%%%%%%%%%%%%%%%%%%%%%%%%%%%%%%%%%%%%%%%%%%%%%%%%%%%%%%%%%%%%%%%%

\section{Experimental Application to Glaucoma Datasets \label{sec:experiment}}

We compare the effectiveness of our proposed 
hierarchical aggregating strategy against a single cluster-based predictor for glaucoma datasets.

\subsection{Problem Setting}

The dataset used in this study is the same as that used by Liang et al. \cite{Liang:2013}.
We divided the dataset into two parts,
a learning set ($90\%$, 977 patients) 
and a test set ($10\%$, 109 patients).
In the learning period,
we constructed experts using clustering methods,
and used them to determine RG-optimal and IR-optimal $\eta$ values.
In the test period,
we predicted the final observation point for each patient.
The prediction errors were evaluated using ten-fold cross-validation.

We set the number of spatial clusters to $K = 40$
and the number of slope clusters to $C = 5$, as reported in \cite{Liang:2013}.
In our dataset, $D = 74$.
Clusters were omitted if they contains fewer than three patients.
LR, SC, and TSLR gave 977, 977, and 38 experts, respectively.
For all calculations,
we used the loss function $l(x,y) := || x - y||_2 / (30\sqrt{D})$.

\subsection{Determination of the Optimal Learning Rate $\eta$}

\begin{figure*}[!tb]
 \centering
 \subfigure{\includegraphics[width = 16.0cm,clip]
{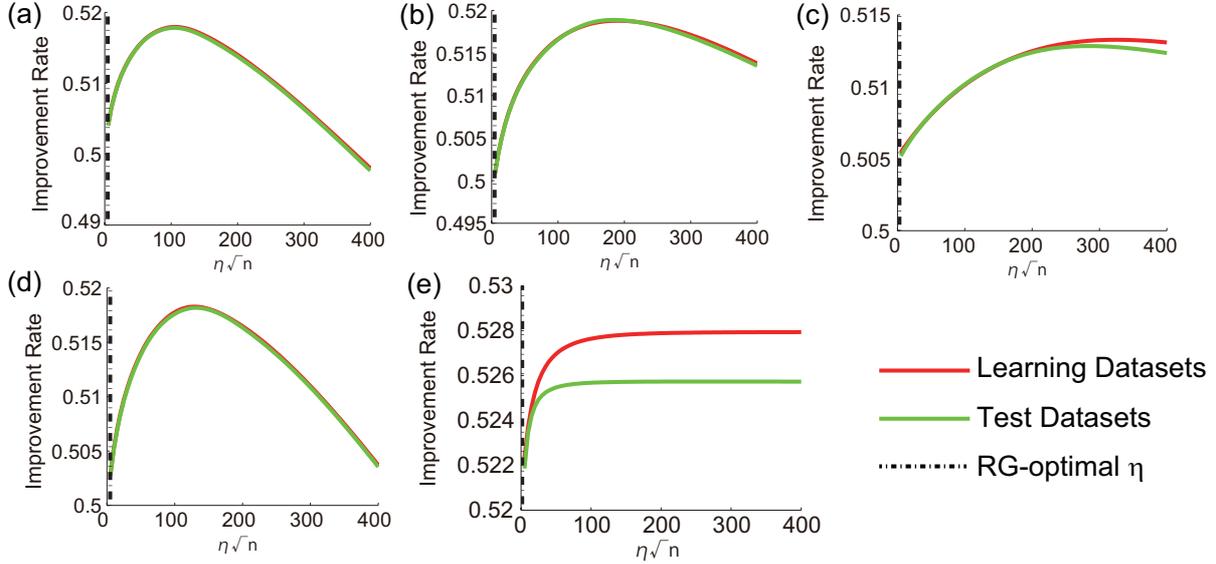}}%
 \caption{
	Dependence of IR on $\eta\sqrt{n}$ for (a) LR, (b) SC, (c) TSLR, 
(d) flat aggregation, and (e) hierarchical aggregation. 
}
\label{fig:obt_eta}
\end{figure*}%

The IR-optimal $\eta$ is obtained 
through ten-fold cross-validation using portions of the learning dataset.
As shown in Fig.~\ref{fig:obt_eta},
IR attained peaks at different values of $\eta\sqrt{n}$.
The IR-optimal $\eta$ generally agrees with the optimal $\eta$ 
obtained using  the test dataset.
However, the RG-optimal $\eta$ are clearly smaller than the optima.

\subsection{Prediction using Experts Generated by a Clustering Method}

\begin{table*}[!t]
\begin{small}
% \begin{scriptsize}
%\begin{tiny}
 \begin{center}
  \caption{IR versus observation point $n$ using IR-optimal $\eta$, RG-optimal $\eta$, the best expert, and the methods of Liang et al. \cite{Liang:2013} based on LR (upper), 
TSLR (middle), and SC (lower) experts. 
IRs corresponding to the best performance are underlines. 
We denote cases in which our method is statistically-significant better than 
Liang et al.'s methods with * ($p < 0.05$) and ** $(p < 0.01)$.}
  \centering
  \label{tbl:TSLRandSC}
  \begin{tabular}{|c|r|r|r|r|r|r|r|r|r|}
  \hline
 LR & $n=2$  & 3 & 4 & 5 & 6 & 7 & 8 & 9 & 10 \\ \hline
 IR-optimal & \ \  \udl{0.852} &  \ \ \udl{0.735} &  \ \ \udl{0.620} &  \ \ \udl{0.518} &  \ \ \udl{0.420} &  \ \ \udl{0.324} &  \ \ \udl{0.242} &  \ \ \udl{0.183} &  \ \ \udl{0.154} \\
 RG-optimal & \ \ 0.852 &  \ \ 0.731 &  \ \ 0.613 &  \ \ 0.506 &  \ \ 0.403 &  \ \ 0.298 &  \ \ 0.214 &  \ \ 0.150 &  \ \ 0.124 \\
 Best expert & 0.695 &  0.569 & 0.461 & 0.376 & 0.296 & 0.222 & 0.158 & 0.116 & 0.0920
\\ \hline
 TSLR & $n=2$  & 3 & 4 & 5 & 6 & 7 & 8 & 9 & 10 \\ \hline
 IR-optimal  &  **\udl{0.853} & **\udl{0.734} &  **\udl{0.618} & **\udl{0.513} & **\udl{0.414} & **\udl{0.314} & **\udl{0.232} & **\udl{0.171} & **\udl{0.144} \\
 RG-optimal  & **0.852 & **0.733 &  \ \ *0.614 &  \ \ 0.506 &  \ \ 0.402 &  \ \ 0.295 &  \ \ 0.209 &  \ \ 0.143 &  \ \ 0.116 \\
 Best expert & 0.834 & 0.709 & 0.590 & 0.486 & 0.390 & 0.292 & 0.212 & 0.154 & 0.131 \\
 Liang et al.  &  \ \ 0.850 &  \ \ 0.728 &  \ \ 0.609 &  \ \ 0.499 &  \ \ 0.395 &  \ \ 0.289 &  \ \ 0.200 &  \ \ 0.136 &  \ \ 0.109 \\ \hline
%%%%%%%%%%%%%%%%%%%%%%%%%%%%%%%%%%%%%%%%%%%%%%%%%%%%%%%%%%%%%%%%%%%%%%%%%%%%%%%%%%%%%
 SC & $n=2$  & 3 & 4 & 5 & 6 & 7 & 8 & 9 & 10 \\ \hline
 IR-optimal  &  **0.850 &  **\udl{0.734} &  **\udl{0.620} &  **\udl{0.519} &  **\udl{0.421} &  **\udl{0.324} &  **\udl{0.243} &  **\udl{0.185} & **\udl{0.156} \\
 RG-optimal  &  **\udl{0.851} & **0.729 &  **0.611 &  **0.502 &  **0.399 &  **0.293 &  **0.209 &  **0.145 &  **0.119 \\
 Best expert & 0.761 & 0.637 & 0.521 & 0.425 & 0.335 & 0.253 & 0.180 & 0.133 & 0.106 \\
 Liang et al.  &  \ \ 0.818 &  \ \ 0.678 &  \ \ 0.554 &  \ \ 0.447 &  \ \ 0.347 &  \ \ 0.256 &  \ \ 0.178 &  \ \ 0.120 &  \ \ 0.0895  \\ \hline
  \end{tabular}
 \end{center}
\end{small}
% \end{scriptsize}
%\end{tiny}
\end{table*}

\begin{figure*}[!t]
 \centering
 \subfigure{\includegraphics[width = 16.0cm,clip]
{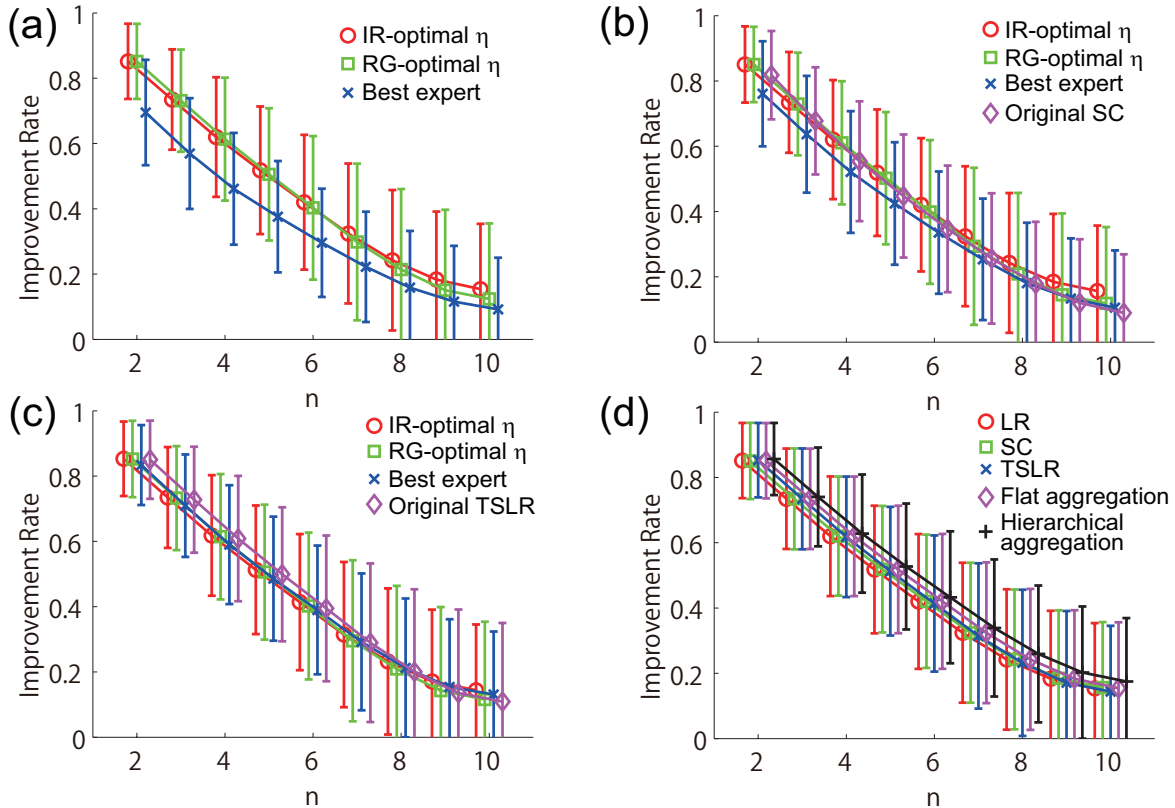}}%
 \caption{
 Dependence of IR on the number of observations $n$ 
for experts generated using (a) LR, (b) SC, and (c) TSLR.
The performance of the patient-wise LR corresponds to the zero horizontal line.
(d) IRs for the five aggregation algorithms with IR-optimal $\eta$ are compared.
Error bars show the standard deviations of $a_i(n)$. 
}
\label{fig:single_expert}
\end{figure*}%

We compared  our aggregating strategies with IR-optimal $\eta$ and RG-optimal $\eta$
with the predictions of the best expert
(i.e., that with the largest weight at the final observation point)
and the methods of Liang et al. \cite{Liang:2013}.
As shown in Figs.~\ref{fig:single_expert}(a)-(c) and Table \ref{tbl:TSLRandSC},
the aggregating strategy with IR-optimal $\eta$ produced a greater IR than that of the
other methods in almost all cases.
We applied a one-sided binomial statistical test to these results
by counting the number of ``wins'' in prediction accuracy among the patients in the test dataset.
The null hypothesis that the prediction ability of our method is the same as that of Liang et al.'s method
was rejected in many cases (shown in Table~\ref{tbl:TSLRandSC} with symbols * $(p < 0.05)$ and ** $(p < 0.01)$, where the p-index is denoted by $p$).
Thus, aggregating cluster information allows for better prediction performance 
than using only one cluster.

Our method with IR-optimal $\eta$ also outperformed the best expert.
This indicates that the clusters do not adequately represent the patient characteristics.
This result also suggests that existing glaucoma prediction methods 
could be enhanced using our aggregating algorithm.

\subsection{Prediction using Hierarchical Aggregating Algorithms}

The hierarchical aggregating strategy with IR-optimal $\eta$ 
achieved better IR values than the flat aggregating strategy and single clustering methods.
In addition,
a one-sided binomial statistical test showed that
the hierarchical approach with IR-optimal $\eta$ was statistically-significantly better 
than the original TSLR and SC in all cases.
These results are summarized in Fig.~\ref{fig:single_expert}(d) and Table~\ref{tbl:all}.

Our proposed aggregating algorithm clearly works very well.
Aggregating algorithms with IR-optimal $\eta$ exhibited 
the best performance in almost all cases,
as shown in Tables~\ref{tbl:TSLRandSC} and \ref{tbl:all}.
Moreover, 
the hierarchical aggregating strategy with IR-optimal $\eta$  achieved the best  performance 
in terms of IRs and the one-sided binomial test ($p < 0.01$) as shown in Fig.~\ref{fig:single_expert}(d).
This implies that the hierarchical method is the best means of 
aggregating the experts generated by clustering-based methods for glaucoma datasets.

\begin{table*}[!t]
% \begin{small}
 \begin{center}
  \caption{IR versus observation point $n$ using IR-optimal $\eta$ and RG-optimal $\eta$.%, and the best expert. 
IRs corresponding to the best performance are underlined.
Statistically-significant results with respect to TSLR  \cite{Liang:2013} are indicated with
* ($p < 0.05$) and ** ($p < 0.01$).
In all cases, the proposed method was significantly better than original SC \cite{Liang:2013} with $p < 0.01$.
}
  \label{tbl:all}
\begin{small}
% \begin{scriptsize}
%\begin{tiny}
  \begin{tabular}{|c|r|r|r|r|r|r|r|r|r|}
  \hline
 Strategies & $n=2$  & 3 & 4 & 5 & 6 & 7 & 8 & 9 & 10 \\ \hline
 Flat  (IR) & *0.851 & **0.734 & **0.620 & **0.518 & **0.420 & **0.324 & **0.243 & **0.184 & **0.154  \\
 Hierarchical  (IR) & **\udl{0.856} & **\udl{0.740} & **\udl{0.628} & **\udl{0.527} & **\udl{0.433} & **\udl{0.339} & **\udl{0.259} & **\udl{0.202} & **\udl{0.175} \\
 Flat  (RG) & 0.851 & **0.731 & *0.612 & 0.504 & 0.401 & 0.297 & *0.212 & 0.148 & 0.122 \\
  Hierarchical  (RG) & **0.853 & **0.733 & **0.615 & **0.506 & 0.403 & 0.298 & *0.212 & 0.148 & 0.121 \\
\hline
  \end{tabular}
\end{small}
% \end{scriptsize}
%\end{tiny}
 \end{center}
% \end{small}
\end{table*}

\subsection{Discussion}

Hierarchical aggregation outperformed the flat aggregating strategy for the following reasons.
If there are a few experts,
those with larger weights (i.e., better experts) in a clustering method will lose 
their influence in the flat aggregating strategy,
because the initial weight becomes small during the flat aggregation process 
over a large number of experts.
Such experts will survive in the hierarchical aggregating strategy, 
because this approach gives equal initial weights to all clustering methods,
regardless of how many experts they include.
In other words, the hierarchical aggregating strategy has a mechanism 
whereby good predictors in a small community (i.e., a clustering-based method with a small number of experts) can thrive.
This works particularly well when the number of experts varies widely across all of the cluster-based methods.
This was the case in our experiments where the three clustering-based methods
had 977, 977, and 38 experts for LR, SC, and TSLR, respectively, 
and TSLR produced a very good predictor (see Table~\ref{tbl:all}).

When the number of experts does not deviate widely,
the advantages of the hierarchical strategy are not so pronounced.
Such a situation might cause the data to be overfitted 
through the repeated aggregation. 
Hence, whether flat or hierarchical aggregation is preferable 
depends on the distribution of the experts in the clustering-based methods,
as well as the nature of the data itself.

%%%%%%%%%%%%%%%%%%%%%%%%%%%%%%%%%%%%%%%%%%%%%%%%%%%%%%%%%%%%%%%%%%%%%%%%%%%%%%%%%%%%%%%%%%%%%%%%%%%%%%%%%%%%%%%%%%%%%%%%%%%%%%%%%%%%%%%%%%%%%%%%%%%%%%
%%%%%%%%%%%%%%%%%%%%%%%%%%%%%%%%%%%%%%%%%%%%%%%%%%%%%%%%%%%%%%%%%%%%%%%%%%%%%%%%%%%%%%%%%%%%%%%%%%%%%%%%%%%%%%%%%%%%%%%%%%%%%%%%%%%%%%%%%%%%%%%%%%%%%%
%%%%%%%%%%%%%%%%%%%%%%%%%%%%%%%%%%%%%%%%%%%%%%%%%%%%%%%%%%%%%%%%%%%%%%%%%%%%%%%%%%%%%%%%%%%%%%%%%%%%%%%%%%%%%%%%%%%%%%%%%%%%%%%%%%%%%%%%%%%%%%%%%%%%%%

\section{Conclusion \label{sec:conclusion}}

In the present study, 
we have developed prediction algorithms 
that aggregate the experts generated by clustering-based methods.
Our aggregation framework has been then applied to the prediction of glaucoma progression.
There are three main differences between our proposed method and the existing approach.
First, we have used an empirically optimized learning rate.
Second, the expert weights are determined by using a batch process.
Third,  
we have examined the hierarchical aggregating strategies 
for multiple clustering-based methods.  
We found that the hierarchical aggregating strategy gives consistently better predictions 
than the traditional patient-wise linear regression, single clustering-based methods, 
the best expert, and the flat aggregation.

The findings reported here may contribute to knowledge discovery and data mining 
by connecting clustering methods and predictions through aggregation algorithms.
Our method for aggregating experts generated several clustering methods
that obtained better prediction results than the original clustering-based method.
This suggests that the prediction accuracy could be further improved 
by embedding additional clustering methods.

From a clinical significance viewpoint,
our method may help to establish good predictions of glaucoma progression.
In addition to its good prediction performance,
our framework of aggregating cluster-based predictors 
may be sufficiently flexible to include other clustering methods and predictors, 
because the algorithm does not assume any specific constraints.
This flexibility will allow clinicians to add novel prediction methods. 
Therefore, our method will contribute to improving the future performance 
of glaucoma predictions.
Moreover, our proposed method is not limited to  glaucoma progress prediction.
Therefore, our proposed aggregating strategies for clustering-based methods 
can be applied in a wide range of areas.

\subsubsection*{Acknowledgments.} 

This work is partially supported by JST-CREST.

\bibliographystyle{abbrv}
%\bibliography{glaucoma}

\end{document}